\documentclass{article}

\usepackage{arxiv}

\usepackage[utf8]{inputenc} 
\usepackage[T1]{fontenc}    
\usepackage{hyperref}       
\usepackage{url}            
\usepackage{booktabs}       
\usepackage{amsfonts}       
\usepackage{adjustbox}
\usepackage{float} 
\usepackage{nicefrac}       
\usepackage{mathtools}
\usepackage{microtype}      
\usepackage{subcaption}
\usepackage{lipsum}		
\usepackage{graphicx}
\usepackage{natbib}
\usepackage{doi}

\title{HAT-CL: A Hard-Attention-to-the-Task PyTorch Library for Continual Learning}

\author{ \href{https://orcid.org/0000-0002-3782-608X}{\includegraphics[scale=0.06]{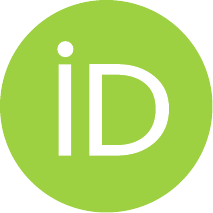}\hspace{1mm}Xiaotian Duan}\
    \\
	Department of Computer Science\\
	The University of Chicago\\
	Chicago, IL 60637 \\
	\texttt{xduan7@uchicago.edu} \\
}

\hypersetup{
pdftitle={HAT-CL},
pdfsubject={q-bio.NC, q-bio.QM},
pdfauthor={David S.~Hippocampus, Elias D.~Striatum},
pdfkeywords={First keyword, Second keyword, More},
}

\begin{document}
\maketitle

\begin{abstract}

Catastrophic forgetting—the phenomenon in which a neural network loses previously obtained knowledge during the learning of new tasks—poses a significant challenge in continual learning. 
The Hard-Attention-to-the-Task (HAT) mechanism has shown potential in mitigating this problem, but its practical implementation has been complicated by issues of usability and compatibility, and a lack of support for existing network reuse.
In this paper, we introduce HAT-CL, a user-friendly, PyTorch-compatible redesign of the HAT mechanism. 
HAT-CL not only automates gradient manipulation but also streamlines the transformation of PyTorch modules into HAT modules. 
It achieves this by providing a comprehensive suite of modules that can be seamlessly integrated into existing architectures.
Additionally, HAT-CL offers ready-to-use HAT networks that are smoothly integrated with the TIMM library. 
Beyond the redesign and reimplementation of HAT, we also introduce novel mask manipulation techniques for HAT, which have consistently shown improvements across various experiments.
Our work paves the way for a broader application of the HAT mechanism, opening up new possibilities in continual learning across diverse models and applications.

\end{abstract}

\keywords{Continual Learning \and PyTorch \and Machine Learning Libraries}

\section{Introduction}

Continual Learning (CL), a paradigm in machine learning, aims to learn from a sequential stream of data, mirroring the human ability to continually acquire, fine-tune, and transfer knowledge throughout life. 
However, artificial neural networks face a significant obstacle in achieving this aim: catastrophic forgetting. 
This phenomenon, wherein the network forgets previously learned tasks while learning new ones, presents a substantial challenge in developing models that can effectively retain and leverage past knowledge.

Various strategies have been proposed to mitigate catastrophic forgetting. 
One such strategy is the Hard-Attention-to-the-Task (HAT) mechanism, proposed by \cite{serrà2018hat}. 
Despite its potential, the practical implementation of HAT is fraught with difficulties. 
The original HAT implementation involved manual gradient modification, which required changing the gradient after backpropagation, for both network parameters and masks. 
Additionally, the implementation of hard attention masks is often tedious and prone to error. 
On top of that, the original HAT demands modifications to both the network structure and the optimization procedure.
In sum, the original HAT implementation does not easily adapt to existing network architectures, limiting its applicability and compatibility.

In this paper, we introduce HAT-CL, a user-friendly, PyTorch-compatible redesign of the HAT mechanism that addresses these issues. 
HAT-CL offers the following advantages over the original implementation:

\begin{itemize}
    \item It encapsulates the hard attention masks within weighted layers, eliminating the need for boilerplate code.
    \item It handles all gradient modifications automatically under the hood, thereby preserving the standard PyTorch forward/backward functions.
    \item It integrates seamlessly with the PyTorch Image Models (timm) library (\cite{wightman2019timm}), enabling easy access to HAT versions of popular models such as ResNet (\cite{he2015resnet}) and ViT (\cite{dosovitskiy2021vit}), with pretrained weights.
    \item It implements better initialization and mask scaling strategies that consistently yield superior results, especailly in smaller networks.
\end{itemize}
The source code for HAT-CL is publicly available at \url{https://github.com/xduan7/hat-cl}.

The rest of this paper is structured as follows: Section \ref{sec:related_work} provides background information and discusses related work. Section \ref{sec:design_and_implementation} details the methodology behind HAT-CL. Section \ref{sec:experiements_and_case_studies} presents our experimental setup and results. We discuss the implications of our findings and future work in Section \ref{sec:future_work}, and finally, conclude in Section \ref{sec:conclusion}.

\section{Related Work}
\label{sec:related_work}

This section revisits the Hard Attention to the Task (HAT) mechanism, as proposed by Serrà et al.~\cite{serrà2018hat}, outlining its strengths and identifying challenges. This discussion provides the necessary background and motivation for our work—HAT-CL.

\subsection{The HAT Mechanism}

Catastrophic forgetting is a significant challenge in continual learning, where a neural network loses previously learned information while learning new tasks. The HAT mechanism, an approach to mitigate this problem, uses attention masks to modulate the contribution of each neuron to specific tasks. This mechanism facilitates efficient allocation of network capacity across multiple tasks, preventing the loss of previously learned tasks when new ones are introduced. 

For a network with \(L\) weighted layers, each task \(t \in [0, 1, ..., T - 1]\) has a set of embeddings \(\mathbf{e}_l^t\), where \(l \in [0, 1, ..., L - 1]\) indexes the layers. The dimensions of these embeddings correspond to the output dimensions of each layer. During the forward pass, the network output is masked by element-wise multiplication between the original layer output \(\mathbf{h}_l\) and the expanded mask \(\mathbf{a}_l^t\):

\begin{equation}
\begin{aligned}
\mathbf{a}_l^t &= \sigma(s \mathbf{e}_l^t) \\
\mathbf{h}'_l &= \mathbf{a}_l^t \odot \mathbf{h}_l
\end{aligned}
\label{eq:hat_fwd}
\end{equation}

Here, \( \sigma \) is the sigmoid function serving as the attention gate, and \(s\) governs the "hardness" of the masks. Smaller scales make the mask embeddings more trainable, while larger scales push the mask towards a binary state, which is desirable for assigning parameters to specific tasks.

The backward pass of the training of task \(t\), requiring careful gradient management, nullifies the gradients of parameters associated with previous tasks. This is achieved by multiplying the layer gradient \(\mathbf{g}_l\) by a factor derived from the previous and current masks:

\begin{equation}
\begin{aligned}
\mathbf{a}_l^{\leq (t - 1)} &= \max \Big( \mathbf{a}_l^{(t - 1)}, \mathbf{a}_l^{\leq (t-2)} \Big) \\
g'^{t}_{l,ij} &= \Big[ 1 - \min \Big( a_{l,i}^{\leq {(t - 1)}}, a_{l-1,i}^{\leq {(t - 1)}} \Big) \Big] g^{t}_{l,ij}
\end{aligned}
\label{eq:hat_bwd}
\end{equation}

These equations (\ref{eq:hat_fwd} and \ref{eq:hat_bwd}) encapsulate the core of the HAT mechanism: isolating parameters for different tasks and nullifying the gradients of parameters associated with previous tasks. However, the training of masks is slow due to smaller gradients. To compensate, we adjust the gradients of the mask embeddings \(q_{l,i}\):

\begin{equation}
q'_{l,i} = \frac{s_{\max} \Big[ \cosh{ \big( s e^t_{l, i}} \big) + 1 \Big]}{s \Big[ \cosh{ \big( e^t_{l, i}} \big) + 1 \Big]} q_{l, i}
\label{eq:hat_comp}
\end{equation}

Here, \(s_{\max}\) is the upper bound of the mask scale. For numerical stability, we clamp the mask embeddings \(e^t_{l,i}\) and the scaled term \(se^t_{l, i}\).

During each training epoch, the scale of masks increases linearly from a really small number to \(s_{\max}\):
\begin{equation}
s = \frac{1}{s_{\max}} + \Big( {s_{\max}} - \frac{1}{s_{\max}}  \Big) \frac{b -1}{B-1}
\label{eq:hat_scale_linear}
\end{equation}
where \(b = 1,2,\dots,B\) is the index of the current batch, and \(B\) is the total number of batches.

Lastly, to prevent overuse of the masks for the current task, we introduce a regularization term:

\begin{equation}
R_t = \sum^{L-1}_{l=0} \max \Bigg( \frac{ \sum_i a_{l,i}^{t} ( 1 - a_{l,i}^{<t} )}{ \sum_i ( 1 - a_{l,i}^{<t} )} - \frac{1}{T} , 0 \Bigg)
\label{eq:hat_reg}
\end{equation}

This term \ref{eq:hat_reg} differs from the original work in two ways: (1) it calculates the regularization per layer and then sums it up, and (2) it incorporates a task quota, such that a task does not get penalized if its mask usage does not exceed the quota.

\subsection{The Implementation of HAT}

While several implementations of HAT exist, such as those presented in \cite{kim2022clom} and \cite{ke2023cl_llm}, they are mostly derived from the original work of Serrà et al.~\cite{serrà2018hat}, and therefore inherit its limitations. Specifically, these implementations still require manual modification of gradients, which significantly restricts the compatibility and reusability of the network.

Implementing a network using the HAT mechanism involves defining a masking function corresponding to Equation \ref{eq:hat_fwd} and a gradient modifier as per Equation \ref{eq:hat_bwd} for each weighted layer. Furthermore, after backpropagation, users need to manually apply the gradient modifier in \ref{eq:hat_bwd}, compensate for the gradient \ref{eq:hat_comp}, and clamp the embeddings. 

This process is intricate and prone to errors. Specifically, the requirement of accessing parameters by their keywords poses a significant risk of introducing errors. Moreover, the complexity of the process impedes the extension of the network as each weighted layer requires careful attention. Implementing large-scale networks such as vision transformers becomes particularly challenging, let alone accounting for the ever-growing variants of transformers.

To the best of our knowledge, no implementation has successfully addressed these challenges, highlighting the need for a more user-friendly and flexible approach, which we aim to provide with HAT-CL.

\section{Design and Implementation}
\label{sec:design_and_implementation}

This section delves into the design principles of the HAT-CL library and some specific implementation details that equip users to either utilize or extend the library to suit their needs. We sequentially explore the payload of the module, basic modules, and implemented networks, detailing their design and functionality.

\begin{figure}[ht]
    \centering
    \begin{subfigure}[b]{0.45\textwidth}
    \begin{flushright}
        \includegraphics[height=9cm,keepaspectratio]{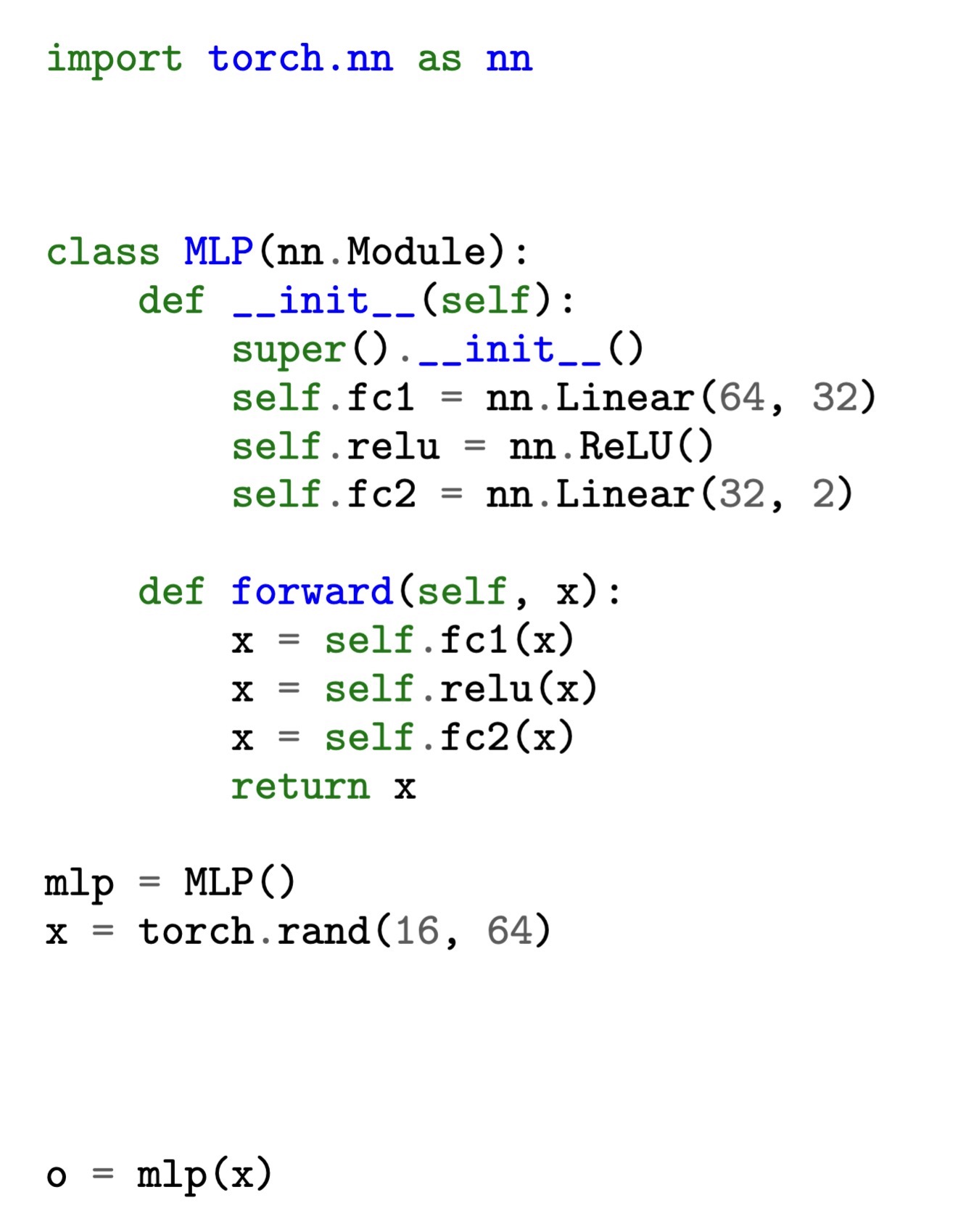}
        \caption{PyTorch Network}
        \label{fig:your_first_label_here}
    \end{flushright}
    \end{subfigure}
    \begin{subfigure}[b]{0.45\textwidth}
    \begin{flushleft}
        \includegraphics[height=9cm,keepaspectratio]{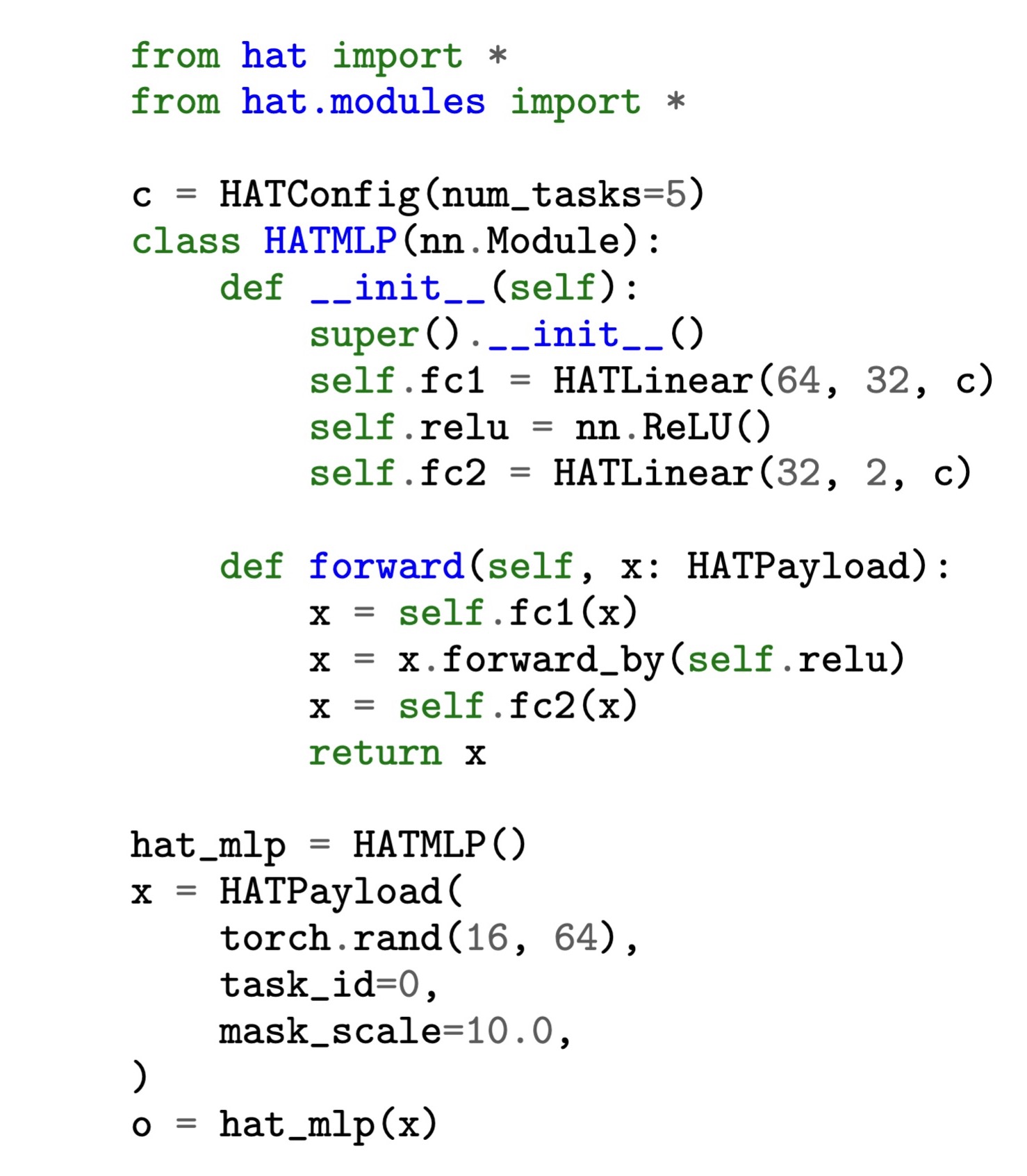}
        \caption{HAT Network}        
    \end{flushleft}
    \end{subfigure}
    \caption{Comparison between a normal PyTorch network and its HAT counterpart}
    \label{fig:networks}
\end{figure}

A simple example that demonstrates the basic usage of HAT-CL is shown in figure \ref{fig:networks}

\subsection{Payload}

At the heart of the HAT-CL library is the \texttt{HATPayload} class, designed to encapsulate tensors alongside relevant HAT information such as the task ID and mask scale. 
This class serves as the input and output for all modules implemented within the library. 
Furthermore, the \texttt{HATPayload} keeps track of the sequence of HAT masks it traverses, thereby implicitly providing the mask ordering crucial for the gradient nullification process as specified in Equation \ref{eq:hat_bwd}.

One of the key design features of \texttt{HATPayload} is its lazy mask application. 
In essence, the HAT mask is only applied when the masked data is accessed. 
This feature proves particularly useful in scenarios where the original data should be preserved to yield the correct output. For instance, when applying layer normalization, input masking can lead to unstable results due to the resultant near-zero values. 
In such cases, layer normalization needs to be applied prior to the HAT mask.

Finally, to ensure flexibility and compatibility with complex network architectures like ResNet and Transformer, \texttt{HATPayload} implements common tensor operations such as \texttt{reshape}, \texttt{permute}, addition, and matrix multiplication. 
These operations are safely applied to the unmasked data in the payload, ensuring the library's usability and robustness in various continual learning contexts.

\subsection{Modules}

The HAT-CL library implements two types of modules

\begin{itemize}
    \item \texttt{HAT} modules: Implement the HAT mechanism (e.g., \texttt{HATLinear} or \texttt{HATConv2d}).
    \item \texttt{TaskIndexed} modules: Dispatch the input to different submodules of the same type based on the task ID (e.g., \texttt{TaskIndexedBatchNorm2d} or \texttt{TaskIndexedLayerNorm}).
\end{itemize}

While they both prevent catastrophic forgetting by isolating parameters, these modules differ in their applicability and function within the HAT framework. 
\texttt{HAT} modules contain weight parameters that require the HAT mechanism to prevent catastrophic forgetting. 
These parameters are shared across tasks to enable knowledge transfer, with updates restricted by the HAT mechanism to prevent interference with learned tasks. 
In contrast, \texttt{TaskIndexed} modules, such as batch normalization layers, do not require the HAT mechanism, as their weight parameters are not meant to be shared across tasks.

\subsubsection{\texttt{HAT} Modules}

The \texttt{HAT} modules pair a base module, such as \texttt{torch.nn.Linear}, with an attention mask that corresponds to the output dimension.
The forward pass of a \texttt{HAT} module takes the masked data from a \texttt{HATPayload}, passes it through the base module, and then applies the necessary operations for the HAT mechanism in \ref{eq:hat_bwd} if the module is in training mode. 
These operations are registered as backward hooks, allowing the HAT mechanism to function without any user interference. 
During the retrieval of the masked data, the unmasked data will pass through the pending attention mask, and during the forward pass, the hook for gradient compensation in \ref{eq:hat_comp} will be registered, along with gradient and value clamping. 
Thus, \texttt{HAT} modules automate the majority of the HAT mechanism, only excluding regularization.

\subsubsection{\texttt{TaskIndexed} Modules}

\texttt{TaskIndexed} modules prevent forgetting by completely isolating parameters using a \texttt{torch.nn.ModuleList}. Essentially, a \texttt{TaskIndexed} module acts as a list of base modules. During the forward pass, the module dispatches the input tensor to the submodule corresponding to the task ID indicated in the payload.

\subsubsection{Other Modules}

Some modules, such as dropout, activation, and pooling layers, don't require special treatment to be used in a HAT network, as they are neither weighted nor influenced by task IDs. 
However, these modules don't accept \texttt{HATPayload} as input. To handle this, we've implemented a class method for \texttt{HATPayload}, \texttt{HATPayload.forward\_by}, which allows the masked data to be passed through layers that don't inherently accept \texttt{HATPayload}.

\subsection{Networks}

In an effort to provide a seamless experience for users and enhance the compatibility of our library, we have designed HAT-CL to integrate smoothly with the PyTorch Image Models (TIMM) library \cite{wightman2019timm}. To this end, we have implemented HAT versions of several popular image networks, replicating the interfaces used in TIMM.

\begin{figure}[ht]
    \centering
    \includegraphics[width=0.8\textwidth]{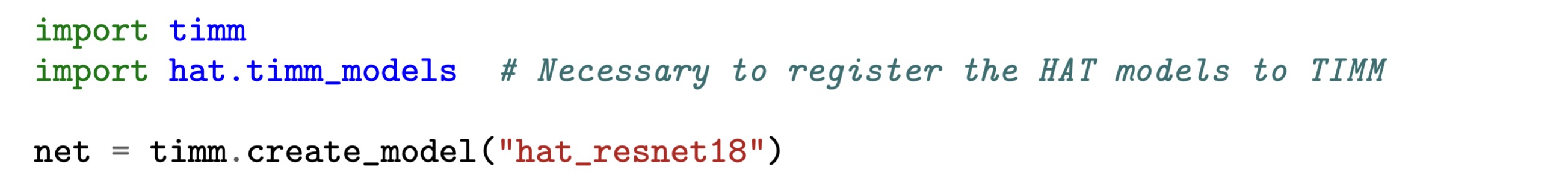}
    \caption{Creating a HAT version of ResNet18 using the \texttt{timm.create\_model} function.}
    \label{fig:timm}
\end{figure}

To distinguish these HAT variants, we append the prefix "\texttt{hat\_}" to the standard network name. For instance, the HAT version of ResNet18 can be accessed using the identifier "\texttt{hat\_resnet18}", as demonstrated in Figure \ref{fig:timm}.



An additional advantage of our HAT network implementation is its support for pretrained weights. For networks such as Vision Transformers (ViTs), pretrained weights can be loaded directly onto the corresponding HAT modules. For \texttt{TaskIndexed} modules, we load the pretrained weights onto all modules within the list. This feature facilitates a faster training process and allows users to leverage pre-existing models for their continual learning tasks.

\section{Experiments and Case Studies}
\label{sec:experiements_and_case_studies}

In this section, we will go over some experiments and case studies that demonstrate the usage of the library other than continual learning, or experiement on different approaches compared to the original implementation.

\subsection{Initialization and Mask Scaling}

In the original HAT implementation, the authors suggested initializing masks using a Gaussian distribution, specifically  \(\mathcal{N}(0, 1)\), which they found to yield superior results.

However, this approach can lead to unintended consequences. Consider a scenario where a feature that should ideally contribute positively to a task happens to be initialized with a negative weight. 
This misalignment leads to the feature being penalized during training, causing the corresponding mask value to near zero. 
This makes it challenging to adjust the weights during training so the feature can contribute positively, as it should.

This problem is particularly noticeable in smaller networks. 
Given the smaller size of their weight matrices, the probability of a feature being suppressed incorrectly is higher. 
This aspect of mask initialization could be a limiting factor in the performance of smaller networks using the HAT mechanism.

In contrast, our implementation, HAT-CL, proposes a different mask initialization and scaling strategy. Instead of a Gaussian distribution, we initialize all mask embeddings with a value of 1. The mask scale then follows a cosine curve within a unit training time, typically an epoch. So different from \ref{eq:hat_scale_linear}, we use the following scaling approach:

\begin{equation}
s = \frac{s_{\max}}{2} \cdot \Big( 1 + \cos{( p * 2 \pi )} \Big)
\label{eq:hat_scale_cosine}
\end{equation}
where \(p\) is a number between \(0\) and \(1\), indicating the progress in the current unit training time.

This approach divides a unit training time into three phases:

\begin{enumerate}
    \item In the first phase, the mask scale is high, restricting the training of the masks. Here, the weights are the primary target of training, ensuring they align correctly before the masks are trained.
    \item In the second phase, the mask scale approaches 0, allowing the masks to train with the aligned weights and reducing the likelihood of the aforementioned problem.
    \item In the final phase, the mask scale increases again, fine-tuning the weights on the nearly binary masks to enhance the final prediction accuracy.
\end{enumerate}

To demonstrate the effectiveness of our approach, we compared it with the original implementation using a simple network with five input features, only three of which are useful. We consider a network to be properly trained when the attention masks reflect the usefulness of the features. We counted the number of batches required to train the network and repeated the process for 100 networks in each case. The average number of batches until training completion is shown in Table \ref{tab:init}.

\begin{table}[H]
	\caption{Sample table title}
	\centering
	\begin{tabular}{cc}
		\toprule
		\cmidrule(r){1-2}
		Strategy          & Avg. Number of Batches\\
		\midrule
		Original          & 338.50                \\
		HAT-CL (ours)     & 25.00                 \\
		\bottomrule
	\end{tabular}
	\label{tab:init}
\end{table}

As shown, HAT-CL significantly reduces the number of batches needed to properly train the network, highlighting the advantages of our mask initialization and scaling strategy.

\subsubsection{Selective Forgetting}

One of the unique features of this library is the ability to selectively forget certain tasks while preserving the knowledge of others. This is possible because the network parameters are associated with specific tasks.

To forget task \(t\), we must scrutinize each module and identify parameters that are solely associated with task \(t\) and not with any others. It may not always be possible to find such parameters, but for larger networks, the chances of not finding them are negligible.

The HAT-CL library provides a utility function, \texttt{forget\_task}, which enables a network to forget a specified task using the method described above. This function returns a dictionary that tracks the number of parameters forgotten during the process.

Table \ref{tab:forget} demonstrates this feature using a ResNet model trained on CIFAR-10 dataset (split into five tasks by class) and then selectively forgetting task 0.

\begin{table}[H]
    \caption{Selective Forgetting of Task 0 in a Continual Learning Scenario}
    \centering
    \begin{tabular}{cccccc}
        \toprule
        \cmidrule(r){1-6}
        & Task 0 Acc. & Task 1 Acc. & Task 2 Acc. & Task 3 Acc. & Task 4 Acc. \\
        \midrule
        After Training on Task 0 & 99.65\% & - & - & - & - \\
        After Training on Task 1 & 99.65\% & 95.05\% & - & - & - \\
        After Training on Task 2 & 99.65\% & 95.05\% & 96.55\% & - & - \\
        After Training on Task 3 & 99.65\% & 95.05\% & 96.55\% & 98.75\% & - \\
        After Training on Task 4 & 99.65\% & 95.05\% & 96.55\% & 98.75\% & 97.80\% \\
        After Forgetting Task 0 & 50.00\% & 95.05\% & 96.55\% & 98.75\% & 97.80\% \\
        \bottomrule
    \end{tabular}
    \label{tab:forget}
\end{table}

In this table, it's evident that after forgetting task 0, the accuracy for task 0 significantly decreases while the accuracies for the other tasks remain the same, demonstrating the successful application of selective forgetting.

\section{Future Work}
\label{sec:future_work}

While our library, HAT-CL, has shown promise in overcoming catastrophic forgetting in continual learning, there are several potential directions for future work. 

One direction could involve testing HAT-CL with a wider range of neural network architectures, including variants of vision transformers and large language models. This would help assess the versatility and adaptability of HAT-CL across different types of networks.

Additionally, future work could explore how HAT-CL could be used in a broader range of tasks and domains. While our current work has focused primarily on image classification tasks, HAT-CL could potentially be used for natural language processing, reinforcement learning, audio signal processing, and other areas of machine learning.

Further optimization of HAT-CL could also be a focus for future work. This could involve improving computational efficiency, extending functionality, or adding new features.

Integration of HAT-CL with other machine learning libraries or frameworks could also be explored. This would make HAT-CL more accessible to a larger audience and allow for easier incorporation into existing machine learning workflows.

Finally, further empirical validation of HAT-CL would be beneficial. Future work could involve testing HAT-CL on a broader range of datasets, tasks, and continual learning scenarios to further validate its performance and applicability.

\section{Conclusion}
\label{sec:conclusion}

In this paper, we introduced HAT-CL, a user-friendly, PyTorch-compatible redesign of the Hard-Attention-to-the-Task (HAT) mechanism for mitigating catastrophic forgetting in continual learning. Our work provides significant improvements over the original HAT implementation, with features such as automated gradient manipulation, simplified transformation of PyTorch modules into HAT modules, and seamless integration with the TIMM library. 

We have shown that HAT-CL is capable of handling large and complex network architectures, providing ready-to-use HAT versions of popular models. Through extensive experimentation, we've demonstrated the effectiveness of our library in maintaining task-specific knowledge and its potential for selective forgetting.

Furthermore, we proposed a novel mask initialization and mask scaling strategy that demonstrates superior results, especially in smaller networks. 

HAT-CL opens up new possibilities for the application of the HAT mechanism across diverse models and applications, making it an invaluable tool for researchers and practitioners in the field of continual learning. We look forward to seeing the community adopt and expand upon our work.

\bibliographystyle{unsrtnat}
\bibliography{references}  






\end{document}